\documentclass[sigconf]{acmart}

\AtBeginDocument{%
  \providecommand\BibTeX{{%
    \normalfont B\kern-0.5em{\scshape i\kern-0.25em b}\kern-0.8em\TeX}}}

\setcopyright{acmcopyright}
\copyrightyear{2018}
\acmYear{2018}
\acmDOI{10.1145/1122445.1122456}

\copyrightyear{2022}
\acmYear{2022}
\setcopyright{acmcopyright}\acmConference[CHI '22]{CHI Conference on Human Factors in Computing Systems}{April 29-May 5, 2022}{New Orleans, LA, USA}
\acmBooktitle{CHI Conference on Human Factors in Computing Systems (CHI '22), April 29-May 5, 2022, New Orleans, LA, USA}
\acmPrice{15.00}
\acmDOI{10.1145/3491102.3501818}
\acmISBN{978-1-4503-9157-3/22/04}

\usepackage{balance}       
\usepackage{graphics}      
\usepackage[T1]{fontenc}   
\usepackage{mathptmx}
\usepackage{color}
\usepackage{booktabs}
\usepackage{textcomp}
\usepackage{eurosym}
\usepackage{ljh}
\usepackage{mathtools}
\usepackage{multicol}

\begin{document}

\title[Roman]{Roman: Making Everyday Objects Robotically Manipulable with 3D-Printable Add-on Mechanisms}

\author{Jiahao Li}
\affiliation{%
  \institution{UCLA HCI Research}
  \country{} 
}
\email{ljhnick@g.ucla.edu}

\author{Alexis A Samoylov}
\affiliation{%
  \institution{UCLA HCI Research}
  \country{} 
}
\email{alexiy.samoylov@gmail.com}

\author{Jeeeun Kim}
\affiliation{%
  \institution{HCIED Lab, Texas A\&M University}
  \country{} 
}
\email{jeeeun.kim@tamu.edu}

\author{Xiang `Anthony' Chen}
\affiliation{%
  \institution{UCLA HCI Research}
  \country{} 
}
\email{xac@ucla.edu}



\begin{abstract}

One important vision of robotics is to provide physical assistance by manipulating different everyday objects, e.g., hand tools, kitchen utensils. However, many objects designed for dexterous hand-control are not easily manipulable by a single robotic arm with a generic parallel gripper. Complementary to existing research on developing grippers and control algorithms, we present Roman, a suite of hardware design and software tool support for robotic engineers to create 3D printable mechanisms attached to everyday handheld objects, making them easier to be manipulated by conventional robotic arms. The Roman hardware comes with a versatile magnetic gripper that can snap on/off handheld objects and drive add-on mechanisms to perform tasks. Roman also provides software support to register and author control programs. To validate our approach, we designed and fabricated Roman mechanisms for 14 everyday objects/tasks presented within a design space and conducted expert interviews with robotic engineers indicating that Roman serves as a practical alternative for enabling robotic manipulation of everyday objects.
\end{abstract}



\begin{teaserfigure}
  \includegraphics[width=\textwidth]{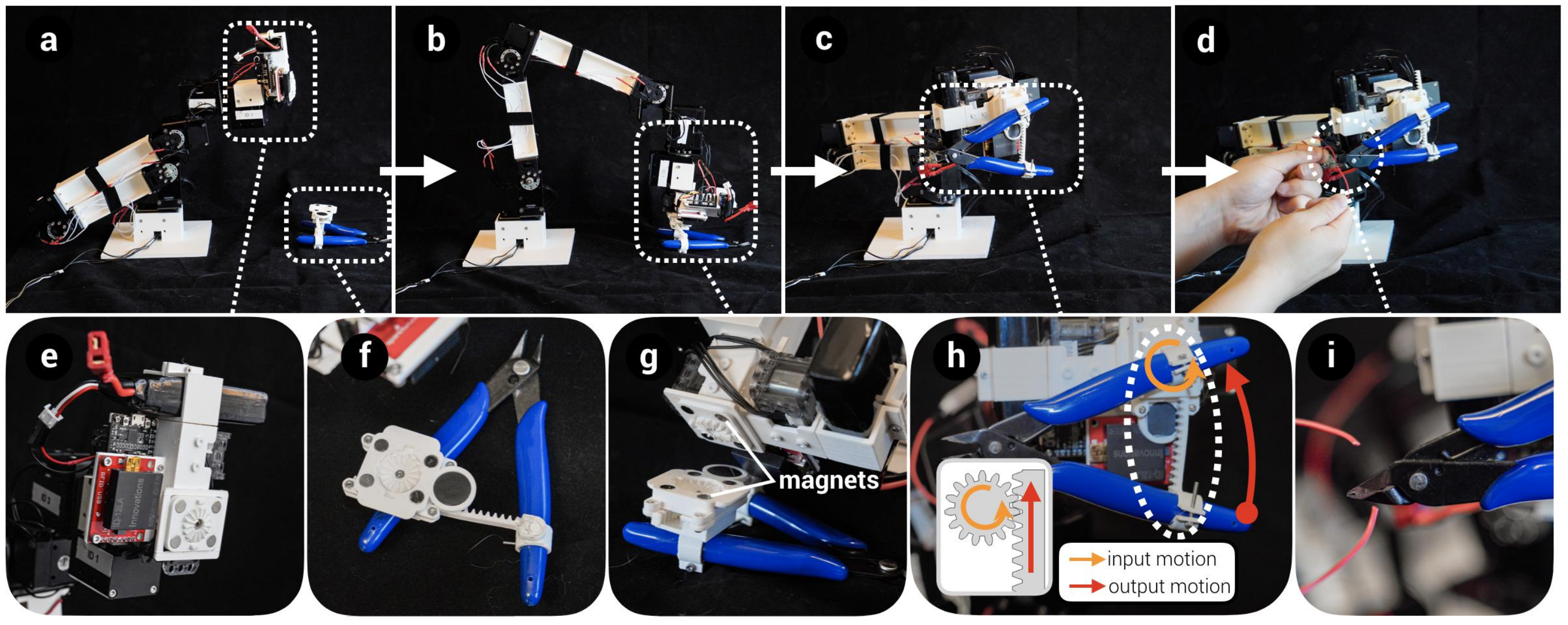}
  \caption{{Roman is a novel robotic design making everyday handheld objects more robotically manipulable, \ie easier to be manipulated by a conventional robotic arm. This figure shows a sequence of a Roman-enabled robotic arm picking up a wire cutter on the desk and performing a wire cutting task collaborating with a human (a-d). Roman provides a magnetic gripper (e) for the robotic arm to easily attach and augment the wire cutter with Roman mechanism (f). Snapping in the mechanism using magnets (g), the gripper can actuate the gear-rack movement on the wire cutter (h) to perform the cutting task by squeezing the cutter's handles (i).}}
  \label{fig:fig1}
\end{teaserfigure}

\maketitle


\section{Introduction}




Various types of robots that inhabit our living spaces, such as vacuum cleaners, robotic toys, and home/office-assistive robots, promise to aid humans with numerous everyday tasks. 
Future robots are expected to provide physical assistance for the elderly at home or help with household chores that require the use of diverse tools and everyday objects \cite{kemp2007challenges}.


Many recent advances in robotic research (primarily on manipulation \cite{wang2019vision, manuelli2019kpam, fischinger2016hobbit, wang2020vision}) aim to solve robotic arms' manipulation problem for niche, fixed tasks, \eg opening medicine bottles, handling cookware, and sorting waste. 
However, creating a \textit{universally dexterous} robotic arm remains challenging as there are a large number of everyday objects that are not robotically manipulable, \ie difficult to manipulate by a consumer-level robotic arm with a generic parallel gripper, especially so for those \textit{dynamic} objects that have multiple mechanically movable parts, \eg a pair of pliers and a spray bottle with a pump pushable toward the body.
To tackle this robotic manipulability challenge, prior work often considers 
the action of the gripper and the manipulator as completely decoupled, which reduces the complexity of task planning and control involved \cite{babin2020mechanisms}. 
To date, researchers have focused on developing manipulation strategies by analyzing the best grasping points of \textit{static} objects \cite{wang2019vision, manuelli2019kpam}. 
However, when it comes to \textit{dynamic} objects such as a spray bottle with a pump pushable toward the body, prior work tends to mimic the pose and force of a human with \textit{multiple} robotic arm \cite{edmonds2017feeling, liu2019mirroring} or dexterous robotic hand \cite{andrychowicz2020learning} rather than tackling the task self-containedly with a {\it single} robotic arm. 
Meanwhile, on the objects' side, augmenting objects with actuable mechanisms is a new approach to enhance their interactivity or functionality (\eg \cite{chen2016reprise, li2019robiot}); however, little has been explored on how to enable robotic arms to better manipulate such augmented objects. 

We present Roman---a suite of hardware design and software tool support for robotic engineers to make everyday handheld objects more {\underline R}{\underline o}botically {\underline m}{\underline a}{\underline n}ipulable by a consumer-grade 6-DoF robotic arm. 

Roman's hardware components 
consist of
\one a library of 3D printable powerless mechanisms that are attached to and can drive different handheld objects to perform specific tasks, \eg squeezing a cutter to cut wires (\fgref{fig1}a$\rightarrow$d);
\two a gripper (\fgref{fig1}e) that uses magnets to securely and automatically attach/detach a robotic arm to/from an object\footnote{Caveat: currently, an object's placement needs to be known for auto-attachment}.
The gripper also contains an RFID-based module to recognize which object the robotic arm is expected to manipulate thus to run the corresponding control program for specific tasks.
%
%

Roman's software component is a user interface for robotic engineers to register and specify the motor input of custom tasks with pre-defined templates and real-time feedback.
A user can rapidly and iteratively author a control program for a Roman-enabled robotic arm to manipulate a handheld object.
Note that Roman does {\it not} reinvent tools for generating 3D models of add-on mechanism, which is already supported (\eg \cite{li2019robiot,10.1145/2984511.2984512}); rather, Roman's software focuses on creating a control program to operate such mechanisms as different handheld objects require unique sequences of action to perform a specific task, which would otherwise be tedious to specify even for robotic experts.



To validate Roman, we first designed and fabricated mechanisms for 14 everyday objects to demonstrate how a generic 6-DoF robotic arm (built with 3D printed links and Dynamxiel XH-540 motors\footnote{https://www.robotis.us/dynamixel-xh540-w150-t/})
can manipulate these objects to perform specific tasks.
Further, we conducted a preliminary interview with four robotic engineers. Participants found Roman's approach complementary to the current research on robotic manipulation within the human environment, which could be beneficial in specific task scenarios such as cooking, electronics assembly, and caring for house plants. 
Provided with the pre-fabricated gripper and attaching mechanisms, all participants were able to use Roman's software module to replicate the wire cutter scenario (\fgref{fig1}).

\subsection{Contributions}
Overall, Roman empowers robotics engineers to use 3D printable powerless mechanisms attached to different objects for enhancing these objects' robotic manipulability with task-specific control program embedded in the mechanisms.
Our specific contributions are as follows.
\begin{itemize}
    \item {\bf Categorization of the robotic manipulability problem in everyday handheld objects ({\S}\ref{sec:design_space})}, including the challenge of speed and range of motion, and the challenge of dexterity when manipulating an dynamic object with its constituent parts (squeezing, twisting, and pumping);
    \item {\bf A library of 3D printable add-on mechanisms able to manipulate both static and dynamic objects  ({\S}\ref{sec:mechanism})} based on variations and combinations of spur gears, bevel gears, gear racks, and pin-in-slot mechanisms, which can be attached to everyday objects to enhance their manipulability by a consumer-grade 6-DoF robotic arm;
    \item {\bf The design of a versatile magnetic gripper ({\S}\ref{sec:gripper})} that can securely attach to the add-on mechanisms on an object, run the control program to drive the mechanism, and detach from the mechanism without outside intervention;
    \item {\bf Software that enables robotic engineers to interactively author a program ({\S}\ref{section:ui})} to control the Roman mechanism-installed hardware to perform object-specific tasks.
\end{itemize}

\subsection{Limitation}

At present, Roman has not achieved total autonomy, as there is no sensing modules to detect where an object is positioned or how it is oriented for pick-up (which is a separate topic well studied in robotics and computer vision \cite{elfes1989using, geiger2013vision}). Currently, in most of our examples, the object is handed off to the robotic arm by a human (\eg kitchen utensils) or the position of the object is manually input to the robotic control program by the human user (\eg the spray bottle example).




\section{Categorizing Challenges of Handheld Objects for Robotic Manipulation}
\label{sec:design_space}


Many everyday objects are not robotically manipulable due to the following challenges:

\begin{itemize} [leftmargin=0.25in]
    \item \textbf{Speed and range of motion challenges} when manipulating an object as a whole.
    For example, objects that require a high speed/frequency of manipulation may exceed the motor's capability at the end-effector, \eg a rapidly rotating whisker at a high speed to whip cream (\fgref{limitation}a).
    Meanwhile, manipulating a screwdriver at certain angles might exceed the reach of a situated robotic arm (\fgref{limitation}b) 
    
    \item \textbf{Dexterity-related challenges} when manipulating an object by its constituent parts. For example, to cut a wire, a robotic arm would first need to grasp the cutter legs then apply a steady force to squeeze the handles (\fgref{limitation}c); however, a generic parallel gripper might find it hard to both securely grasp {\it and} squeeze the cutter.
    Bi-manual manipulation is even harder, \eg opening a jar lid (\fgref{limitation}d) is almost impossible for a single robotic arm and requires either a holding stand or assistance from another robotic arm or a human hand.
\end{itemize}

\fgw{limitation}{limitation}{1}{A human hand manipulating a variety of objects that present challenges for robotic arms such as using a whisk (a), rotating a screwdriver (b), squeezing a wire cutter (c), and opening a jar lid (d).}

\fgw{range_of_motion}{range_of_motion}{1}{Why manipulating a screwdriver might require range of motion beyond a robotic arm's capability: (a) normal working grasp/distance; (b) when the distance increases, the grasp changes and the end-effector can no longer rotate around the screwdriver's axis; (c) in some edge cases, the manipulation might be interfered by the physical surroundings, \eg the ground.}



Our goal is to make everyday {\bf handheld objects} manipulable by a consumer-grade generic 6-DoF robotic arm.
To achieve this, we first need to understand \textit{what} objects are currently inaccessible and \textit{how} they are inaccessible for robotic manipulation.
Everyday objects are mostly designed to be manipulated by humans. With flexible fingers and the coordination of four limbs, humans are able to manipulate a wide range of objects with different manipulation complexities, from picking up a cup to playing a piano. 
In contrast, there exist a large number of everyday objects that are not manipulable by generic robotic arms due to the fact that a general-purpose gripper has limitations in performing different types of grasping and manipulation tasks \cite{babin2020mechanisms}.
As shown in \fgref{designspace}, we consider two categories of such objects: those manipulated as a whole (\eg raising/lowering a knife, rotating a screwdriver) {\it vs}. those manipulated by constituent parts (\eg squeezing a pair of pliers' handles, twisting a pepper grinder cap against the bottle, pushing the pump towards the body of a bottle of hand sanitizer). 

\subsection{Objects Manipulated as a Whole (Static Objects)}

As shown in \fgref{designspace}A-B, objects in this category require a `grasp \& move' type of manipulation as the body of the object moves as a whole while performing tasks, \eg the whole knife moves vertically while chopping vegetables (linear motion), the whole egg beater rotates when mixing eggs (rotational motion). 
Although in theory such objects can be manipulated by a robotic arm---by first grasping it using a gripper and then performing manipulation by a series of joint rotations to create the movement, there remain two limitations that prevent the robotic arm from manipulating such objects to perform tasks as well as humans: Speed and Range of motion.

\subsubsection{Speed limitation} 
A robotic arm mostly manipulates objects at its end-effector, which generates large loads on the arm due to the moment of inertia. Therefore, tasks such as knife chopping (\fgref{designspace}A1), which require a fast periodic motion at the end-effector may exceed the robotic arm's capability. 

\subsubsection{Range of motion limitation}
Using a conventional robotic gripper, some object manipulation requires a certain type of grasp, \eg rotating a screwdriver (\fgref{range_of_motion}a).
However, when the object is further way, even though it is still within reach of the robotic arm, the grasp is different and as such, it is no longer able to afford the same manipulation (\fgref{range_of_motion}b).
In other words, grasping the tool at a certain angle might render the manipulation impossible because reaching that angle would already constrain the robotic arm's joint rotation, thus limiting how it can perform subsequent manipulation (\fgref{range_of_motion}b).
In some edge cases, the manipulation might be interfered with by the physical surroundings, \eg the ground (\fgref{range_of_motion}c).
In robotic terms, under such circumstances, the robotic arm is said to be outside of its {\it dexterous space} \cite{lia1988dexterous} when having to perform the manipulation at certain angles.


\fgw{designspace}{designspace}{1}{Everyday handheld objects are often not manipulable by a generic robotic arm with a common parallel gripper: when manipulating an object as a whole, speed and range of motion are two main limitations and; when manipulating an object by its constituent parts, the dexterity required to both grasp and perform a range of manipulation (squeezing/releasing, twisting and pumping) is the main challenge.}


\subsection{Objects Manipulated by Constituent Parts (Dynamic Objects)}

In contrast to objects manipulated as a whole, there exist objects made up of and manipulated by their movable parts, as shown in \fgref{designspace}C-E.
Manipulating such objects tends to be more difficult for a robotic arm, which needs to first grasp the object stably and then perform a dexterous manipulation like human hand(s), \eg grasping a pepper grinder and then twisting the grinding cap (\fgref{designspace}D2). 
We summarize the following three types of manipulation that makes objects not manipulable by generic robotic arms, each of which can be either one- or bi-directional.

\subsubsection{Squeezing}
In order to manipulate objects that require `squeezing' (\eg a wire cutter in \fgref{designspace}C2), a robotic arm (\eg with a common parallel gripper) would have to pick two points on the squeezing handles for a firm grasp, and then apply a steady force to squeeze the object. However, as the best grasping points for performing the squeezing manipulation (\eg near the tip of the handle ) are usually the furthest away from the center of gravity, this makes the grasp unstable and slippery.


\subsubsection{Twisting}
The twisting force applied to objects \eg a door knob (\fgref{designspace}D1) would produce a rotational torque on the robotic arm itself after securing the grasp, which may cause the whole system to become unstable. This is also the most common failure in the \textit{DARPA Robotic Challenge} \cite{krotkov2017darpa}.

\subsubsection{Pumping}
Different from the above, after grasping an object, the robotic arm needs an additional contact surface to perform the pumping task (\eg to press the hand sanitizer in \fgref{designspace}E1), which makes it nearly impossible to perform by a common gripper with a parallel design.

Finally, note that some objects (\eg pepper grinder, pump) would require bi-manual manipulation. In other words, a {\it single} robotic arm, even as dexterous as a human hand, would find it challenging to manipulate these objects.







\section{Related work}



\subsection{Augmenting Generic Interfaces and Everyday Objects}
Roman can be thought of as enabling two types of augmentation: \one augmenting a generic robotic arm so that it can attach to and manipulate various handheld objects and \two augmenting everyday objects so that they become manipulable by the robotic arm.
As such, we first review prior work on these two aspects.

\subsubsection{Augmenting generic physical interfaces}
Existing work has explored various designs for extending the functionality of generic interfaces. For example, HERMITS extended the capability of self-propelled tangible user interfaces (TUIs) by designing mechanical shell add-ons for TUIs to dock and drive dynamic interactions \cite{nakagaki2020hermits}. Katakura \etal introduced a novel way of augmenting a 3D printer head into a robotic manipulator with the mechanical attachments printed by the printer itself \cite{katakura20193d, katakura2018printmotion}. Other researchers augmented a pin-based display by the conversion of the mechanical motion of pins with passive mechanisms to enrich the pins' dynamic interaction \cite{nakagaki2020trans, schoessler2015kinetic}. Davidoff \etal designed mechanical actuators based on the Lego MindStorm toolkit to operate on switches that were intended for humans only \cite{davidoff2011mechanical}. In comparison, Roman's mechanisms augment the physical interface of a robotic arm, \ie the end-effector, with a versatile magnetic gripper with a focus on assisting the robotic arm to manipulate different everyday objects that would otherwise be difficult to manipulate. 

\subsubsection{Augmenting everyday objects}
Another focus of previous research is extending the capability of everyday objects. Reprise makes handheld objects easy-to-manipulate by people with disabilities by generating adaptions attached to the object \cite{chen2016reprise}. Medley enhanced the material properties of 3D printed objects by embedding different reusable materials into the model \cite{chen2018medley}. Romeo extended the default functionality of everyday objects by embedding a transformable robotic arm into the 3D model of the object \cite{li2020romeo}. RetroFab augmented the interactivity of physical interface by adding an enclosure consisting of mechanical and electrical components that could automate physical controls \cite{ramakers2016retrofab}, which was later extended by Robiot in automating dynamic everyday objects, \eg adjusting a lamp's joint angle \cite{li2019robiot}. 

RetroFab and Robiot employed {\it active} mechanism (\ie with motors onboard), 
which inevitably makes the augmented object bulky and expensive to scale to the numerous everyday objects in the environment. 
In contrast, Roman utilized a passively actuable mechanism attached to an object with only mechanical components to be manipulated by a generic robotic arm. 

Since Roman proposes a different complementary approach for robotic arms to grasp and manipulate everyday objects, below we will review related work in robotics research, focusing on gripper design and manipulation.

\subsection{Relationship to Robotic Research}
\subsubsection{Gripper design}
In Robotics, grippers are the most common type of end-effector and an important medium for robots to interact with the real world. The design of robotic grippers has been extensively studied in academia and industry by researchers and practitioners to explore designs of different types of robotic end effectors, \eg grippers for pick and place, tight grasping, and more. There are two major strategies in designing a gripper: for general purposes or for a specific task. Researchers have explored different types of general purpose robotic grippers including linkage-based parallel mechanism robotic grippers \cite{hassan2014design, kocabas2009gripper} and compliant underactuated robotic grippers \cite{lee2020twister, ciullo2020novel, liu2018optimal, 7222634}. 
Meanwhile, exploring how to configure task-specific grippers is an emergent topic that has gained recent attention. Feix \etal provided a taxonomy to categorize the potential tasks and the corresponding design features of the robotic gripper \cite{7243327}. 
Researchers have designed grippers for different shapes of target objects \cite{nakayama2019designing}, \eg for picking objects lying on flat surfaces \cite{babin2018picking, babin2019stable}, for assembly tasks \cite{xu2021selecting}, or for operating a heavy load \cite{takaki2006100g}. 

Furthermore, the design of robotic grippers should not only consider the geometry of the objects, but also the interaction with the environment and the kinetostatic properties of the grippers \cite{babin2020mechanisms}. Different factors are considered in previous research such as dynamic loads \cite{negrello2020benchmarking} and active surfaces after grasping \cite{6491289, golan2020jamming}. These are closely related to the tasks conducted by robotic grippers in daily life interacting with everyday tools, which are summarized by the robotic grasping and manipulation challenge \cite{engeberg2018robotic}. 
Roman uniquely combines both strategies by implementing a general purpose versatile gripper that could grasp different objects and object-specific mechanisms that could be driven by the gripper to achieve a dynamic manipulation of everyday handheld objects.

Another important gripper-related topic is to create `tool changers' that adapt to the different shapes and sizes of everyday objects. 
There are different types of tool changers for a robotic arm: automatic tool changers that are electro-mechanically actuated \cite{gyimothy2011experimental} or passive mechanisms actuated by a host robot \cite{berenstein2018open, pettinger2019passive}. 
Similar to our method, researchers also proposed the design of mechanical tool for robots to grasp different daily objects with modular two-finger gripper but only focused on the grasping of objects with different sizes \cite{hu2019designing}. 
Adding to that, our method could solve a more complex manipulation problem of manipulating tools with movable parts (\fgref{designspace}), which is almost unattainable using a two-finger type end effector. 
Furthermore, since Roman provides customized mechanisms for different objects that can be driven by the same custom gripper, our method provides a less complex method to program the motion of the arm than solutions that incorporate tool changers.

\subsubsection{Manipulation}
Besides innovating the gripper design, researchers have also investigated 
methods of control and task planning that program a given robotic arm to perform tasks in human environment.
Some research focused on developing a control strategy based on perception such as analyzing the geometry of the objects to obtain the best grasping point \cite{wang2019vision, manuelli2019kpam} or imitating human operation to open medicine bottles \cite{edmonds2017feeling, liu2019mirroring}. 
Some developed new algorithms and system designs for specific contexts such as grasping flat objects \cite{sarantopoulos2018grasping}. 
At the application level, robots can assist people with daily living tasks which range from fetching a mug \cite{fischinger2016hobbit} to taking medications \cite{kostavelis2018ramcip}. 
In contrast to the above approaches of enhancing perception and control, Roman aims for a different and complementary goal of achieving manipulation by augmenting everyday objects to be more manipulable by a generic robotic arm.

Different from traditional robotic research, a concept of \textit{dexterous manipulation} was first defined by Okamura \etal in which a robotic gripper moves objects from one configuration to another \cite{okamura2000overview}, \eg adjusting the angle of a phone in the hand.
Such a concept is still being actively studied by researchers in scenarios such as robotic in-hand manipulation \cite{ueda2010multifingered, andrychowicz2020learning, zimmermann2020multi}. However, such manipulation requires precise control of the forces and motions of fingered or specialized robotic hands and therefore cannot be accomplished by conventional robotic grippers. Roman, as a complementary solution to dexterous manipulation, enables a conventional robotic arm to connect with a versatile magnetic gripper to grasp and manipulate everyday objects with dexterity enabled by object-specific mechanisms.

\section{Examples: Roman Makes Handheld Objects Robotically Manipulable}
%

We showcase a series of examples where Roman mechanisms attached to handheld objects make them more manipulable by a consumer-grade 6-DoF robotic arm (equipped with a Roman gripper). There are two major types of application scenarios where the objects need to be robotically manipulable: \one A human collaborates with a Roman-equipped robotic arm, \eg the human would hand an object over to the robotic arm where it would then be picked up and manipulated, in tasks that the robot is expected to manipulate a lot of different handheld objects, \eg making a scrambled egg (\fgref{whole2}de, \fgref{oil}, etc) and \two a Roman-equipped robotic arm takes the place of human in performing repetitive tasks by manipulating a hard-to-manipulate object, \eg a spray bottle.
We focus on demonstrating the wide capabilities of Roman in enabling object manipulation, while discussing the technical details of the gripper design and the mechanism generation method later in subsequent sections.

\subsection{Manipulating Objects as a Whole}

\subsubsection{Increasing speed}
As mentioned above, some objects may require a high speed of manipulation that exceeds the motor's capability at the end-effector. \fgref{whole1} and \fgref{whole2} showcase several different examples where Roman enables a high speed of manipulation: by attaching a pin-in-slot mechanism, the Roman gripper can drive a knife to perform repetitive high-speed vertical motion, \eg for chopping scallions (\fgref{whole1}ab), 
or to mimic a human's tapping motion on the spice bottle to spread enough white pepper (\fgref{whole1}cd). With a spur-gear mechanism, the Roman gripper enables a high speed of rotation of the whisk to mix an egg (\fgref{whole2}de).



\fg{whole1}{as_whole_1}{1}{Configuration of the pin-in-slot mechanism (a) to produce periodic up and down motion with a knife (b)
and an alternative configuration of the pin-in-slot mechanism (c) to produce periodic side to side motion for spreading peppers from a spice bottle (d).
}

\subsubsection{Expanding range of motion}
Roman helps expand the range of motion while manipulating specific objects. For example, grasping a screwdriver at certain distances/angles may prevent a conventional robotic arm from performing the rotating manipulation (\fgref{range_of_motion}). Therefore, Roman adopts a bevel gear mechanism that could change the rotational axis of the input, expanding the range of motion of the screwdriver's manipulation (\fgref{whole2}a-c). Similarly, by attaching the whisk at a small angle (as opposed to being perpendicular) when fastened to the output spur gear, we can expand the range of rotational motion at its end (\fgref{whole2}de). 

\fg{whole2}{as_whole_2}{1}{Configuration of the bevel gear mechanism to produce rotational motion at an angle with a screwdriver (a), allowing for two attachment configurations (b, c); the spur gear mechanism with a reverse reduction gear ratio to produce higher rotational speeds with a whisk (d) and tilting the whisk by a small angle further expands the range of motion at its end (e).}

\fgh{spray}{spray}{1}{Configuration of the gear rack mechanism to produce linear motion (a) in a single direction to squeeze the handle of a spray bottle (b).}




\subsection{Manipulating Objects by Constituent Parts}

\subsubsection{Enabling squeeze \& release manipulation}
As shown in \fgref{fig1}, the wire cutter augmented with a gear-rack mechanism can perform the task of cutting a wire by squeezing the handles (\fgref{fig1}h). 
Similarly, a gear-rack is used for the spray bottle and thus a robotic arm could fetch the bottle and water the flower automatically (\fgref{spray}). 
As the spray bottle only requires one-directional manipulation, the mechanism only needs to squeeze in one direction before releasing the handle to return to its original position.
\fg{chop}{chop}{1}{Configuration of the spur gear mechanism to produce rotational motion to squeeze together the tips of chopsticks (a), and a practical demonstration of the chopsticks being manipulated with the mechanism attached (b).}
The chopsticks 
require a bi-directional squeezing and releasing manipulation to pick and place food and a spur gear 
mechanism is used for the manipulation (\fgref{chop}).
A gear-rack mechanism is also utilized for the can opener in order to perform the squeezing and rotating manipulation to pierce the peanut butter can (\fgref{can_opener}dg). Specifically, a ratchet design (\fgref{can_opener}ef) is used to `lock' the squeezing of the handles (\fgref{can_opener}h), which then enables the gripper to detach from the handles, attach to the cutting part (\fgref{can_opener}c), and drive the rotation of the handle to open the can (\fgref{can_opener}h). 


\fgw{can_opener}{can_opener}{0.8}{Combination of two separate mechanisms with a reduction gear ratio to produce high torque: a spur gear to cut around the can (c), and a gear rack to pierce the can (also using spur gears to increase torque) (d). A ratchet mechanism (f) is used to maintain the position of the piercing mechanism (\ie keep the handles squeezed), which allows the gripper to switch to the other mechanism (c) for cutting the can open.
}



\subsubsection{Enabling twist manipulation}
Following the manipulation of piercing the can, the robotic arm is able to twist the handle continuously to open the can by using a spur gear mechanism (\fgref{can_opener}ch). Since both the squeezing and the twisting manipulation of the can opener require a relatively large strength to manipulate, a gear box with ratio of 9:1 is adopted to increase the output torque applied to the target object (\fgref{gearbox}d). 
Using a pepper grinder is important in a series of cooking tasks (\eg making an omelette). With a bevel gear mechanism, the robotic arm can twist the grinder repetitively to sprinkle pepper on the eggs (\fgref{jar_and_grinder}cd). 
A robotic arm can also collaborate with a human in a cooking task by opening the lid of a starch jar with a bevel gear mechanism (\fgref{jar_and_grinder}ab).
Roman also enables a robotic arm to open the door by twisting the door knob with a spur gear mechanism on it (\fgref{knob}).

\fgh{knob}{knob}{1}{Configuration of the spur gear mechanism with a reduction gear ratio to produce rotational motion with high torque in order to rotate the door knob (a), and a practical demonstration of the door knob being twisted with the mechanism attached (b, c).}

\fgh{jar_and_grinder}{jar_and_grinder}{1}{Two examples of the bevel gear mechanism to produce rotational motion at an angle to unscrew the lid (a, b) or to rotate the pepper grinder (c, d).}

\subsubsection{Enabling pump manipulation}
With a gear-rack mechanism, an oil spray can be manipulated by a robotic arm to help human cook (\fgref{oil}).
A robotic arm can fetch a bottle of hand sanitizer augmented with a gear-rack mechanism and pump it when the user approaches it (\fgref{sanitizer}). 
To enable a repetitive manipulation of the balloon pump, a gear-rack mechanism is also adopted to perform a bi-directional manipulation (\fgref{pump}).

\fgh{sanitizer}{sanitizer}{1}{Configuration of the gear rack mechanism to produce bi-directional linear motion in order to squeeze a bottle of hand sanitizer (a), and a practical demonstration of the bottle of hand sanitizer being squeezed with the mechanism attached (b, c).}

\fgh{pump}{pump}{1}{Configuration of the gear rack mechanism to produce bi-directional linear motion in order to actuate a pump (a), and a practical demonstration of the pump being actuated with the mechanism attached (b, c).}

\fgh{oil}{oil}{1}{Configuration of the gear rack mechanism to produce linear motion in order to depress the spray button (a), and a practical demonstration of the button being squeezed with the mechanism attached (b, c).}








\section{System Overview of Roman}





\fgw{overall}{overall_structure}{1}{Overall hardware structure of Roman}

Roman is an all-in-one solution to make everyday objects manipulable by generic robotic arms and includes both hardware and software support:
\begin{itemize}
    \item {\bf Hardware modules} consist of (as shown in \fgref{overall})
        \begin{itemize}
            \item {\bf {\S}\ref{sec:gripper}}: A modular magnetic gripper that can attach to or detach from an object's add-on mechanism, recognize the object to retrieve the corresponding control program, and transfer the driving force from the robotic arm's motor to the mechanism to execute the object-specific manipulation;
            \item {\bf {\S}\ref{sec:mechanism}}: 3D-printable powerless mechanisms (spur gear, bevel gear, gear-rack, and pin-in-slot) attached to the object which enables objects to be manipulated as a whole or by their constituent parts. The mechanisms are easy to remove/assemble using screws.
        \end{itemize}
    \item {\bf {\S}\ref{section:ui}}: {\bf Software module} is a tool for robotic engineers to interactively specify custom motion profiles for manipulating a specific object (\eg amplitudes of a signal over time to be sent to the motor that drives the mechanisms to squeeze a wire cutter).
    We specifically focus on authoring motion profiles, which, to the best of our knowledge, is unsupported by prior work; meanwhile, the task of generating the 3D models of mechanisms---based on the type of motion and an object' geometry---can be supported by existing tools \cite{li2020romeo, li2019robiot, chen2016reprise, chen2016making}.
\end{itemize}



\section{Hardware Implementation}
\subsection{A Magnetic Gripper to Attach to, Recognize, and Transfer Motion to an Object's Add-on Mechanism}

\label{sec:gripper}

The magnetic gripper serves as the intermediary between the robotic arm with two main functionalities: 1) attaching to and detaching from the passively actuable mechanism on the target objects and 2) driving the mechanisms on the target objects to perform the manipulation. Further, the gripper also contains an RFID reader for recognizing which object it is attached to and running the corresponding control program.

\fgw{gripper}{gripper}{1}{Exploded view of the Roman gripper.}
\fgw{mesh_and_detach}{mesh_and_detach}{1}{Meshing (a) and detaching (b) operations of Roman gripper.}



\subsubsection{Attaching and detaching mechanisms:} The gripper uses four neodymium magnets (\fgref{gripper}D) to generate the magnetic force for attaching to the mechanism on the target objects. The four neodymium magnets could generate a pull force (the vertical force required to pull the magnets from a steel surface) of 11.2lbs in total, which is sufficient to securely attach to common everyday handheld objects. Furthermore, the strong magnetic force enables the gripper to attach to objects when it is within approximately 1cm of them, which increases the fault tolerance of the robotic arm's manipulation (\eg a low cost robotic arm\footnote{PincherX 150 Robotic Arm: https://www.trossenrobotics.com/pincherx-150-robot-arm.aspx} may have accuracy larger than 5mm). 
To further strengthen the connection, we designed a pin structure (\fgref{gripper}L) to counter the lateral force generated during the actuation of the mechanisms. 

On the other hand, a stronger connection means larger force required for the detachment. 
To provide auto-detachment of the mechanisms,
Roman employs a Dynamixel XL-320 motor\footnote{https://www.robotis.us/dynamixel-xl-320/} (\fgref{gripper}E) and a gear-rack mechanism (\fgref{gripper}F/O), which could transfer the rotational motion of the motor into linear motion, for the auto-detachment (\fgref{mesh_and_detach}b).
While the XL-320 motor can generate a maximal torque of 0.39 N$\cdot$M, we designed the gear of the gear-rack mechanism to have a radius of 5mm which enables it to generate 17.52lbs of force on the rack for detachment (\fgref{gripper}O).

\subsubsection{Motion transmission:} The gripper also serves as the actuator of the mechanisms on target objects. Same as the detachment, Roman employs an XL-320 motor (\fgref{gripper}B) and a pair of crown gears (\fgref{gripper}C/I). The crown gear could transmit the rotational motion of the motor to the mechanism with auto-alignment of the teeth (\fgref{mesh_and_detach}a). On the side of the mechanism, modular driving gears (\eg gears or gear-rack mechanism, \fgref{gripper}N) can be assembled with the female crown gear to actuate different types of mechanisms.

\subsubsection{Communicating with target object} 
Roman employs NodeMCU ESP-12 module\footnote{https://www.nodemcu.com/} for communicating with the web server (\fgref{gripper}H). An RFID reader is used for recognizing different objects (\fgref{gripper}G) and each mechanism on the object comes with an RFID tag (\fgref{gripper}K) whose ID is associated with a user-defined control program to perform a specific task.
Fasteners can also be customized and 3D printed to fit different robotic arms using screws (\fgref{gripper}P). 
Finally, the gripper can be powered by an additional 7.4v LiPo battery. The whole system weighs 110g without the battery, which makes it possible to be installed and used on any generic robotic arm. 

\label{section:mechanism}

\subsection{Mechanisms for Manipulating Handheld Objects with Different Motion Profiles}
\label{sec:mechanism}
We first discuss the mechanism design on the objects' side that transfers the rotary input from the motor into task-specific motion profile for different objects.

Objects require different motion profiles in order to perform their tasks. The motion profile is defined as the required output motion for the objects in order to perform an object-specific task. For example, objects in the \textit{squeezing} category may require a curved motion profile (\eg the legs of the wire cutter move in an arc trajectory for a cutting task \fgref{fig1}); objects in the \textit{twisting} category require a rotary motion profile (\eg rotating the door knob to open the door \fgref{knob}) and objects in the \textit{pumping} category require a linear motion profile (\eg linear pushing motion for the balloon pump Fig \fgref{pump}). 

To address this, Roman adopts four basic types of mechanical mechanisms: spur gears, bevel gears, gear-and-rack and pin-in-slot mechanisms (\fgref{mechanisms}). These four types of mechanisms are able to transfer the rotary motion from the motor into different motion profiles.

\fgw{mechanisms}{four_mechanisms}{1}{Overview of the four types of mechanisms used to achieve target motions: spur gears for rotational motion (a), bevel gears for rotational motion at an angle (b), gear rack for linear motion (c), and a pin-in-slot mechanism to achieve periodic motion (d).}
\subsubsection{Spur gears} 
Spur gears are a mechanism in which multiple gears mesh together to transmit the rotary motion from one shaft to another (\fgref{mechanisms}a). Therefore this mechanism can transfer the rotary motion from the motor input into a rotary motion profile.

\textbf{Rotary to rotary} \ While the output motion has the same rotary motion as the input, the gear-pair mechanism can translate the rotary axis to a parallel position, which enables the mechanism to be anchored to a fixed point on the object while manipulating the object. As shown in 
\fgref{knob}, instead of being directly anchored to the door knob, which may get in the way of people using it, the mechanism for driving the knob can be shifted to a position where it does not interfere with regular use of the knob. 

\textbf{Speed and strength}
\fgw{gearbox}{gearbox}{1}{Breakdown of the configuration of the gearbox, showing how to produce higher speeds at the expense of torque (a) and to produce higher torque at the expense of speed (b), as well as the physical configuration of the gears (c) and the design of the gearbox (d).}
Besides the translation of the rotational axis, Roman leverages the property of reduction in gears to generate higher speed or strength than is typically available from a motor. Gear reduction is an arrangement of gears in which an input speed can be raised at the expense of torque, or the output torque can be raised at the expense of speed as explained in the following equations:
  \begin{equation*}
      \text{Speed}_{\text{output}} = \frac{r_{\text{input}}}{r_{\text{output}}} \times \text{Speed}_{\text{input}}
  \end{equation*}
  \begin{equation*}
      \text{Torque}_{\text{output}} = \frac{r_{\text{output}}}{r_{\text{input}}} \times \text{Torque}_{\text{input}}
  \end{equation*} 
With a proper selection of the gears, the mechanism can generate much larger torque than the motor's base ability (\fgref{gearbox}). 
In the meantime, with a reversed reduction gear, the mechanism can generate higher rotational speed than the motor's typical maximum speed. The whisk is an example of leveraging reversed reduction gears to enable the whisk to rotate at a high speed to beat the egg (\fgref{whole2}de). Such reduction gears can also be combined with other mechanisms depending on the task (\eg the can-opener adopts a reduction gear box over its gear-rack mechanism to generate enough force to pierce the can as shown in \fgref{can_opener}d). 


\subsubsection{Bevel gears}
Similar to the spur gears, the output of the bevel gears is also rotary motion. Unlike spur gears, bevel gears change the orientation of the rotational axis, which enables robotic arms to execute a twisting task from different angles (\fgref{mechanisms}b).

\textbf{Rotary to rotary} \ Bevel gears can generate a rotary motion profile for target objects. With the property of being able to change the orientation of the rotational axis, bevel gears enable robotic arms to operate twisting tasks with a space-efficient solution. For example, in order to perform a lid opening task, 
a spur gears design would make the mechanisms bulky as the driving gear will protrude from the jar lid. 
In contrast, A bevel gears design rotates the protruding gear so that it reduces the overall volume occupied by the entire mechanism (\fgref{jar_and_grinder}).

\subsubsection{Gear-rack}
The gear-rack mechanisms are utilized to convert the rotary motion from the gear into the linear motion of the rack (\fgref{mechanisms}c). As the gear-rack mechanism generates a linear motion profile, this mechanism can be used by the objects in the \textit{grasp \& pump} category. Further, the gear-rack mechanism can also be utilized for generating a curved motion profile, which will be discussed below. 

\textbf{Rotary to linear} \ While the gear-rack mechanism converts the rotary motion from the gear into the linear motion of the rack, it could help the robotic arm to manipulate objects that require pumping (\fgref{designspace}E). Designers of the mechanism could adjust the output velocity and force by modifying the size of the gear based on the equation:
\begin{equation*}
    \text{Speed}_{\text{output}} = \text{rpm}_{\text{motor}} \times r
\end{equation*}
\begin{equation*}
    \text{Force}_{\text{output}} = \frac{\text{torque}_{\text{motor}}}{r}
\end{equation*} where $r$ represents the radius of the gear attached to the motor. 
 For example, hand sanitizer is an example that requires large torque to squeeze out sanitizer \fgref{sanitizer}. 

\textbf{Rotary to curved} \ Besides the linear motion profile, the gear-rack is also able to generate a curved motion profile for squeezing objects such as the squeezing leg of can opener (\fgref{can_opener}). 
Roman leverages the compliant property of the PLA material to design racks that can bend themselves to adapt to the curved motion (\fgref{can_opener}e). 
As a result, the fastener on the rack (normally attached with the other movable part of the object) would require an active joint in order for the rack to rotate relatively, without which the curved motion may generate a large offset at the tip of the rack and break the mechanism.  


\textbf{Bi-directional manipulation} \ The gear-rack mechanism may deal with objects requiring either one or bi-directional manipulation. For objects with only one-directional manipulation such as the oil spray (\fgref{oil}), a simple bar is sufficient for generating a one-directional force. However, objects that require bi-directional manipulation such as the balloon pump, require an additional structure fastened to the part to enable motion in two directions (\fgref{pump}b). 

\textbf{Single-direction constraints} \ Roman also provides a single direction constraint using the gear-rack mechanism. For example, in order to perform the task of piercing the can and then opening it, the mechanism needs to lock the squeezing mechanism for the opener to continue to pierce through the surface of the can. To achieve this, we employ a design inspired by the ratchet gear, where the rack can move freely in one direction while being prevented from moving in the opposite direction unless a pin is depressed, which releases the mechanism (\fgref{can_opener}d).

\subsubsection{Pin-in-slot}
A pin-in-slot is a mechanism where a pin-joint moves along or parallel to a slide-joint (\fgref{mechanisms}d). The pin-joint receive the rotary input from the motor and transfer the motion to the slide-joint.
By fixing one side of the slide-joint, the pin-in-slot mechanism can generate a single-sided motion profile (\fgref{mechanisms}d bottom) while a double-sided motion profile can also be generated by placing the slide-joint inside of another slider joint (\fgref{mechanisms}d top). 

The spice bottle is an example of using the pin-in-slot mechanism to mimic the human shaking action by generating a single-sided motion profile (\fgref{whole1}cd) and the knife is an example of performing the chopping task by generating a double-sided motion profile (\fgref{whole1}ab). 

\fgw{periodic}{pin-in-slot-periodic}{0.8}{Demonstration of the two types of periodic motion that could be accomplished with a pin-in-slot mechanism: angular (a), and up and down (b).}

\textbf{Periodic motion} \ Besides generating linear and curved motion profiles, the pin-in-slot mechanism is good at generating a periodic motion. As shown in \fgref{periodic}, the continuous rotation of the pin-joint can generate a double-sided or single-sided periodic motion at the slide-joint, \eg for continuously shaking a spice bottle (\fgref{whole1}cd). While it is possible for other mechanisms to generate a periodic motion with periodically changed control input, the control signal may experience data loss when changing intensely in short periods. Differently, the periodic motion generated by the pin-in-slot mechanism is easier to control because it relies on the stability of the mechanism components.

\section{Software Implementation: A Tool for Robotic Engineers to Specify Custom Motion Profiles to Manipulate Object-Specific Mechanisms}
\label{section:ui}

In this section, we illustrate the workflow of a user authoring the control program for a task given some pre-generated mechanisms to manipulate a target object. 
We assume that the users are robotic engineers with 
a mechanical or robotic background and the mechanism generation is supported elsewhere by tools similar to \cite{li2019robiot, chen2016reprise, ha2018computational}.
We exclusively focus on the less-supported part where the users author control programs for different object-specific tasks.
Roman's front end is written in JavaScript and the back end is written in Python. Roman communicates with the ESP8266 module on the NodeMCU and stores the motion profile through a Python-based web server. The RFID reader on the gripper reads the ID of different RFID chips on the objects and accesses the corresponding motion profile via the NodeMCU.

\subsection{Custom Motion Profiles}
As Roman transfers the rotary motion input from the motor to a customized output motion profile, it requires further specification of the motion profiles in order to perform an object-specific task. For example, to design a task for manipulating the wire cutter while collaborating with a human user (\fgref{fig1}), the wire cutter is expected to 
\one move to the desired position for the user to hold the wire for cutting, 
\two squeeze the handles about half-way and hold the position in order for the user to align and adjust the wire, 
\three squeeze fully to cut the wire, and
\four return the handles to the initial configuration. 
While step \one is conducted by the movement of the robotic arm, the rest are done by Roman hardware and require a custom design of the motion profile. More specifically, the customizability of the motion profiles amounts to specifying the amplitudes \textit{u} of the control signal over time, which corresponds to the rotational speed of the motor output. The aforementioned task for the wire cutter has a motion profile as the control signal shown in \fgref{wirecutter_control}.  

\fgw{ui}{user_interface}{1}{The Roman user interface used for selecting motion templates, authoring control programs, and uploading them to the robotic arm.}

\fgw{wirecutter_control}{wirecutter_control}{1}{A sample motion profile used to control the wire cutter in order to cut a wire.}






\subsection{User Interface}
As shown in \fgref{ui}, to facilitate the design of such custom motion profiles, the Roman software provides a user interface for interactively specifying the custom motion profiles by adding or adjusting the key points in the graph of a control signal (\fgref{ui}c). The \textit{u} for the y-axis is a unitless value ranged from -1 to 1 which represents the rotational speed of the motor relative to the maximum speed under the current load (with negative values corresponding to rotation in the opposite direction). The x-axis represents time and the user has the ability to modify the total time the motor is running at a certain speed by adjusting the length of each period of the control signal. With the combination of different control signals, the user could design a custom motion profile for a object-specific task.

\subsubsection{Motion Profile Templates and Interactive Graph Editing} 
Roman provides different ways to interact with the graph of the control signal. As shown in \fgref{ui}a and d, Roman provides four types of motion profile templates for a user to adapt to their needs, including endless rotation (\eg whisk), periodic motion (\eg knife), one-way motion (\eg jar lid), and two-way motion (\eg wire cutter).
By clicking on the corresponding template, the control signal is imported into and visualized on the interactive graph and the user can further adjust the motion profiles. 

To start editing the graph, a user simply double-clicks anywhere on the graph and then adjusts the position of an existing point by dragging it. The user can also adjust the range of the x-axis (\ie increase/decrease time) by clicking on the +/- signs (\fgref{ui}b), which either adds or removes 1 second.

\subsubsection{Real Time Testing}
There is uncertainty in manipulating real-world objects, \eg the manipulation of a wire cutter (the motor's rotational speed) might need to vary with different wires.
As such, Roman enables a real-time testing mode (\fgref{ui}b) that creates a feedback loop in which the user could test and adjust the custom motion profile to see how well the manipulation is performed on the target object. The user could also check the box of `continuous' to specify a repetitive motion.  
%
%
Once a user has tested and is satisfied with the motion profile, they can click on the `save' button to store the motion profile to the target object. 


\section{Validation with Expert interview}

We conducted an expert interview as an initial step to evaluate Roman. The purpose of this interview is to evaluate the usefulness and practicality of Roman in helping robotic engineers achieve object manipulation tasks, as well as to gather feedback and further suggestions from experts in the field of robotics.

\subsection{Participants \& Procedure}

Four participants (all male with an average age of 28.25) with expert knowledge in the domain of robotics and mechanism design were invited for interviews. Three of them are Ph.D. students in the Mechanical Engineering Department with a focus on the research of robotic design and control (P1-P3). Among these, two had prior experience in developing robotic manipulators as a product in a robotic start-up (P1 and P2). The other participant is a post-doc researcher in Electrical Engineering Department with a Ph.D. degree in mechanical engineering and his research focuses on the computational design for origami robots (P4). Although no participants worked in the exact area of robotic manipulation in a human environment, all were familiar with the concept of mechanism design and control of robotic arm manipulation. 


Each of the participants was interviewed in-person. First, they were explained the goal and function of Roman hardware and one author showed them demos of using the Roman gripper to manipulate five different objects (wire cutter, hand sanitizer, jar lid, whisk and spice can). 
These objects covered the four types of mechanisms we used in Roman as well as the four motion profile templates in the user interface. Afterwards, they were introduced to the Roman software and were asked to replicate the custom motion profiles of the wire cutter (which has the most complicated motion profile as shown in \fgref{wirecutter_control}). The motion profiles designed by the participants were test by performing a cutting task (similar to \fgref{fig1}) of an AWG 18 wire. The participants were able to design the profile and cut the provided wire physically in only 5 minutes (excluding the time needed to become familiar with interacting with the user interface)
We then conducted a semi-structured interview, mainly to collect their feedback on the capability and practicality of Roman and potential directions for future research. The entire study lasted on average 25 minutes per participant. 

\subsection{Results and Feedback}

All the participants reacted positively to Roman and gave feedback on both the usefulness and practicality of Roman. 


\subsubsection{Roman was considered complementary to current approaches in robotic manipulation in human environment}
All the participants considered Roman to be a complement to the current research in the robotic manipulation in situations that require robotic arms in human environment. 
P2 stated that most of the existing robotic arms have limited functions or they required complicated programming of the control to use different types of tools. He thought that Roman provided a simple solution as the object + mechanism can be pre-designed as a product and delivered to customers. 
From the perspective of a user, P2 was initially concerned about the practically of possessing a robotic arm in daily life, which was later answered by himself when he noted that the Roman solution could make a robotic arm adapt to different objects in a task-heavy scenario like cooking. 
P1 was concerned about the pre-design of the custom motion profiles to be time-consuming, but he also pointed out that in the application scenarios such as cooking, the number of target objects is limited and therefore it is acceptable. 
P3 considered Roman usable, but he was more concerned about the necessity of the usage of robotic arm in such scenarios versus designing more easy-to-manipulate objects for human (\eg designing a jar that can be opened by pressing a button instead of adding mechanism to be manipulable by robotic arm). 
P4 mentioned that it makes more sense to have a strong and functional robotic hand to manipulate different objects without altering them but he agreed that the artificial hand solution is costly (a powerful enough robotic hand normally costs more than \$10k) and Roman could be a low-cost solution.

\subsubsection{Roman's mechanism designs and custom motion profiles were considered replicable by robotic experts}
When asked about whether they are able to replicate the hardware design of Roman's five examples\footnote{Note that participants did not actually replicate those examples, considering the time-consuming process of modeling, printing and assembly.}, all the participants agreed and were able to understand the rationale behind the design of each mechanism. P1 thought that it would be easy for a person with fundamental knowledge of mechanical design to replicate all of the examples. P2 also pointed out that with the explanation of different mechanisms as in the previous section ({\S}\ref{section:mechanism}) as a guideline, it is a lot easier to select and design the mechanisms for specific objects. P4 stated that it is easy for robotic experts to replicate the examples but it would be great if a parametric design tool is provided for novice users.

All the participants were able to replicate the custom motion profiles for the wire cutter to perform a wire cutting task. All the participants could understand the differences between each motion profile template. P2 also valued the provided templates because, with the templates, he could know the overall movement of the mechanism and would only need to adjust a few parameters. 

\subsubsection{Roman's hardware could be further improved} The participants suggested that the hardware components can be improved to make the overall structure smaller and more durable including developing new types of mechanism (making the mechanisms more versatile), replacing the neodymium magnets with electromagnets (making the mechanisms smaller) and printing the custom mechanism in carbon (making the mechanisms stronger).

\subsection{Summary}
Results of the expert interview indicated that Roman could be a potential solution for robotic manipulation to assist human with daily tasks in certain scenarios. Also, it was considered practical for people with a robotic or mechanical engineering background to design and replicate the objects with mechanisms using Roman. The participants also recommended that a tool for novice users to custom such mechanism for existing objects would be valuable.
\section{Limitations, Trade-Offs, and future work}

\subsection{Performance testing}
Roman's main contribution is the idea of adding mechanisms to enhance objects' manipulability, which complements existing approaches focused on gripper design and manipulation algorithms.
As such, we chose to validate Roman by creating various examples to show the idea's wide applicability and by interviewing four robotic experts to obtain their initial feedbacks and reactions that would set the scene for future work.
As next steps, future work that uses Roman on a specific type of objects/tasks (\eg manipulating a wire cutter) should define metrics for success of each task (\eg a wire is completely cut in one trial), control pertinent variables (\eg the thickness of wires), and employ repeated measures to obtain such metrics.

\subsection{Mechanisms interfering with normal use}
As Roman attaches 3D printed mechanism onto the existing object, some of the mechanisms may interfere with the normal use of the objects. For example, the spur gear mechanism on the chopsticks obstruct the way of a human would use them (\fgref{chop}) and the bulky mechanism on the can opener also makes it hard for human to manipulate (\fgref{can_opener}). While some of the mechanism can be easily disassembled, \eg the rack of the hand sanitizer can be easily pulled out (\fgref{sanitizer}), future work can focus on making the mechanisms modular and easy to disassemble such as LEGO MindStorm \cite{klassner2003lego}. It is also possible to embed such function during the product design stage that aims at making the overall object + mechanism manipulable by both humans and robotic arms.

\subsection{The need to incorporate sensing \& perception}
Roman focuses on enhancing manipulability by enabling human collaborating with Roman-equipped robotic arm or a Roman-equipped robotic arm to perform tasks independently. However, the latter scenario assumes that the object's position and orientation are known, which is a trade-off of not integrating sensing modules in the current design.
Given the plethora of work on sensing and perception (\eg \cite{saxena2008robotic}), future work can add such modules to Roman's hardware components, which are expected to work independently and complementarily to the current set-up.

\subsection{Scale of the task}
Currently, Roman only focuses on tasks driven by the motor of the gripper while movements made by the other joints of the robotic arm have to be manually configured. 
As such, a Roman mechanism only affords manipulation in a limited space and cannot enable large-scale tasks, \eg holding a spatular to make stir-fry.
Future work may extend Roman to include software support involving the entire robotic arm, \eg linking the action of the robotic arm with the actuation of the mechanism. With that, the Roman hardware could collaborate with the robotic arm to achieve more complicated tasks \eg scooping ice cream (enabled by Roman mechanisms) and distributing it into different locations (enabled by the rest of the robotic arm) automatically.

\subsection{Trade-off between torque and the size/complexity of mechanisms}
Currently, Roman enables the manipulation of objects by generating motions using a low-cost motor and 3D printed mechanisms. As a result, Roman is limited to supporting objects that do not require a lot of torque. 
While Roman provides a partial solution by adding a gearbox in between the gripper and the object to increase the torque (see the can opener example \fgref{can_opener}), there is a trade-off between the size of the mechanism and the maximum torque the gripper could generate. For example, manipulating a hedge trimmer (too large) or a sealed jar (too tight) will require rather bulky mechanisms unrealistic to be attached to the object. Substituting the motors in the current design with stronger models could partially solve the problem. Alternatively, there exist research opportunities to solve this problem with improved mechanisms, \eg using more durable materials such as metal linkages or cables to increase the generated force.

\subsection{Manipulating objects with multiple movable parts or multiple consecutive manipulations}
Some objects are articulated with multiple movable parts, \eg multi-functional pliers, Swiss army knife, flexible selfie rods, and Rubik's cubes.
Technically, Roman's mechanisms can extend to more parts by enabling one additional motion at a time, yet, practically, too many mechanisms might not be fittingly added to an object and might even interfere with one another. Future work could focus on involving different materials (cable, metal, carbon) to make the mechanism smaller while functional. 

Other objects might require consecutive manipulations to perform a task, \eg a corkscrew requires a twisting motion on the handle to penetrate a wine cork and a subsequent squeezing motion on the two arms, followed by pulling out the same handle.
Currently Roman needs to provide separate mechanisms for each motion, which could result in the overall mechanisms too bulky. To address this problem, future work could incorporate interactive trajectory design into the mechanism design of Roman, similar to the approach in \cite{coros2013computational} that would need only one custom mechanism to accomplish multiple consecutive manipulations.






\subsection{Opportunities to support novice users}
As mentioned by the robotic engineers in the interview, while it is easy for people with knowledge of robotics to replicate examples in Roman or design mechanism for new examples, novice users might struggle to design mechanism and understand which mechanism to use and how to specify the control signals since Roman does not provide 3D modelling support. 
One possible idea for future work is to instrument sensors on an object similar to the current Roman mechanisms. Then a user can demonstrate how they would manipulate an object, which can be captured by such sensors (\eg inertial measurement units or reflective markers + external optical tracking) and incorporate the algorithm in Robiot \cite{li2019robiot} which could generate the mechanism and control program automatically.
In this way, Roman would be more usable to novice users.

\subsection{Areas of hardware improvement}
The hardware of Roman can be improved by:
\one switching to a half-duplex enabled microcontroller board that could enable the control of the mechanism to be self-adaptive to different loading conditions; \two replacing the neodymium magnets with electromagnets that can be programmatically controlled for attachment/detachment, thus dispensing with the additional motor currently used for detachment.







\bibliographystyle{ACM-Reference-Format}
\bibliography{acmart,references,general,xacpubs}

\end{document}